%% file: main.tex
\documentclass{article}
\usepackage[utf8]{inputenc}
\usepackage[numbers]{natbib}

\usepackage{authblk}
\usepackage{setspace}
\usepackage[margin=1.25in]{geometry}
\usepackage{graphicx}
\graphicspath{ {./figures/} }
\usepackage{subcaption}
\usepackage{amsmath}
\usepackage{amssymb}
\usepackage{amsthm}
\usepackage{lineno}
\usepackage{comment}
\usepackage{enumitem}
\usepackage{bm}
\usepackage{xcolor}

\theoremstyle{definition}
\newtheorem{definition}{Definition}

\newtheorem{property}{Property}

\newcommand{\func}[1]{\textbf{\textcolor{teal}{#1}}}  
\newcommand{\baseC}[1]{\bf{\textcolor{violet}{#1}}}  
\newcommand{\ttime}[1]{\textbf{\textcolor{brown}{#1}}}  

\title{UAMM: Price-oracle based Automated Market Maker}

\author[]{Daniel Jiwoong Im}
\author[]{Alexander Kondratskiy}
\author[]{Vincent Harvey}
\author[]{Hsuan-Wei Fu}

\affil[]{UBET SPORTS\\ubetsports.io\\general@ubetsports.io}

\date{}

\begin{document}

\maketitle

\begin{abstract}
    Automated market makers (AMMs) are pricing mechanisms utilized by decentralized exchanges (DEX). Traditional AMM approaches are constrained by pricing solely based on their own liquidity pool, without consideration of external markets or risk management for liquidity providers. In this paper, we propose a new approach known as UBET AMM (UAMM), which calculates prices by considering external market prices and the impermanent loss of the liquidity pool. Despite relying on external market prices, our method maintains the desired properties of a constant product curve when computing slippages. The key element of UAMM is determining the appropriate slippage amount based on the desired target balance, which encourages the liquidity pool to minimize impermanent loss. We demonstrate that our approach eliminates arbitrage opportunities when external market prices are efficient.
\end{abstract}


\input{intro}
\input{preliminary}

\input{add_refund}

\input{swap_function}

\input{spontaneous_price}
\input{conclusion}

\bibliographystyle{plain}
\bibliography{main}

\input{appendix}

\end{document}

%% file: intro.tex
\section{Introduction}

A Decentralized Exchange (DEX) is a marketplace for trading digital currencies that operates on a blockchain network in a decentralized manner. The popularity of DEX has grown significantly due to its decentralized nature, accessibility, and ability to provide liquidity provision. It consists of pools of funds where the funds are gathered by liquidity providers (LPs) and utilized to provide liquidity to traders. A core component of DEX is an automated market maker (AMM) that algorithmically determines the price of digital currencies without requiring centralized market makers. Users trade directly against pools of funds based on the price given by the AMM instead of relying on a centralized orderbook. For each trade between a pair of quote and base currencies, the liquidity pool balances get updated, where the quantity of the quote currency becomes less, and the base currency becomes more from trade, adjusting the price for each digital currency. LPs earn passive income from the fees generated from these trades.

AMMs use deterministic mathematical formulas to establish the relationship between the quantities of digital currencies in liquidity pools and their corresponding prices. The two most commonly used AMMs are constant function market makers that preserve the relation between the pair of currencies through a constant function concerning the liquidity pool balances \cite{Angeris2021Analysis, Bartoletti2021}.
The most well-known ones are Uniswap \cite{Adams2020UniswapVC} and Curve \cite{Egorov2021}.
Their formulation ensures that the price reflects the supply and demand of the digital currencies, adjusting proportionally to maintain the constant function of the pool as more traders participate in the market.

Unfortunately, existing AMMs that determine prices based on liquidity pool balances, including the constant function market makers, suffer from a major drawback. First and foremost, these AMMs create arbitrage opportunities for traders, resulting in significant losses for LPs. Given the zero-sum game nature of such opportunities, whenever someone capitalizes on arbitrage profits, someone else will inevitably incur losses, and unfortunately, it is typically LPs who bear the brunt as they take the opposite side of the trades. The two primary reasons for these arbitrage opportunities are mispricing and price changes in the currencies. First, AMMs determine prices solely based on internal market information, i.e., the liquidity pool balance, without considering external market information. Consequently, the price of the external market may differ from that of the internal market, exposing LPs to trade at a less favourable price than the fair market price. Moreover, AMMs exhibit slower and delayed adaptation to price changes. It requires someone to seize an arbitrage opportunity resulting from price changes for the AMM to adjust its prices accordingly. This becomes especially problematic when the market does not revert to a stable price or is characterized by high volatility.

In this paper, we propose a new algorithm called \textit{UBET AMM} (UAMM) that addresses the aforementioned issues by eliminating arbitrage opportunities. UAMM calculates prices by considering both external and internal market prices. Despite relying on external market prices, our method maintains the desired properties of a constant product curve when computing slippage. The key element of UAMM is determining the appropriate slippage amount based on the desired target balance, which encourages the liquidity pool to minimize impermanent loss. Therefore, we decompose the price into an estimated fair price and the slippage, estimating each component separately. We demonstrate that our approach eliminates arbitrage opportunities when external market prices are efficient.

In summary, UAMM effectively manages risks for LPs while enabling them to earn passive yields. Unlike Uniswap V3 \cite{Adams2021UniswapVC}, LPs using UAMM need not worry about managing their impermanent loss. We define three transactions: add, remove, and swap, and thoroughly discuss the properties of these UAMM transactions that induce the desired behaviours for an AMM. Moreover, certain properties allow UAMM to eliminate arbitrage opportunities under the assumption that the market price aligns with the fair price.

%% file: preliminary.tex
\section{Preliminary}

We list the notations and variable names for the paper and follow the general notations from \cite{Bartoletti2021}.
\subsection*{Token Notations}
\begin{itemize}
    \item Let $\mathbb{T}$ be the set of {\em fungible atomic tokens} that consists of native or application-specific tokens like ERC20. The fungible atomic token implies that tokens are tradeable and broken down into smaller units. 
    \item Let $\baseC{\tau_0} \in \mathbb{T}$ be the base currency. 
    \item Let $\tau_1, \tau_2, \cdots \tau_K \in \mathbb{T}$ be the tokens of interests for an application. We group them as a unordered tuple of $K$ distinct atomic tokens $(\tau_1, \tau_2, \cdots \tau_K) \in \bigcup^K \mathbb{T}$ where the distinctive atomic token implies that $\tau_i \neq \tau j$ for $i \neq j$. For simplicity, we call them {\em K minted tokens}.
    \item Let $d\tau_i$ be the amount of a token type $\tau_i$. E.g., if base currency $\baseC{\tau_0} = USDC$, then $d\baseC{\tau_0}$ units of USDC.
\end{itemize}

\subsection{States}\label{sec:state}
\noindent The UAMM states vary over time through interacting with users' transactions. The state consists of the liquidity pool balances, fair and spontaneous prices of tokens, total value and supply of the pool, and total received collateral. 
\begin{itemize}
    \item $R\tau_1, R\tau_2, \cdots, R\tau_K$ denote the liquidity pool balances.
    \item $f\tau_1, f\tau_2, \cdots, f\tau_K$ be the fair prices of $K$ minted tokens w.r.t base currency $\baseC{\tau_0}$ given by the oracle.
    \item $p\tau_1, p\tau_2, \cdots, p\tau_K$ be the spontaneous token prices w.r.t base currency $\baseC{\tau_0}$ given by the UAMM.
    \item $TS$ denotes the total supply of liquidity pool share tokens $s_{lp}$. 
    \item $TB$ be the total investment balance in terms of base currency $\baseC{\tau_0}$ (see Definition~\ref{def:tb}).
    \item $TV$ denotes the total value of the liquidity pool (see Definition~\ref{def:tv}).
\end{itemize}
\noindent We denote the state of UAMM as $\Gamma$. 
Most of the time we drop subscript $\ttime{t}$ because it does not require a time dependency.
However, we sometimes explicitly add subscript $\ttime{t}$ to the variable in order to denote the time. For example, $TB_{\ttime{t}}$  and $f\tau_{1,\ttime{t}}$ refers to total investment balance and price of tokens $\tau_1$ at time $\ttime{t}$ respectively.

\subsection{Transactions}
\noindent UAMM transaction from one state to another through user transaction $A$, $\Gamma \xrightarrow{A}{} \Gamma^\prime$. Here is a list of transactions:
\begin{itemize}
    \item $\func{Add}(d{\tau_1}, d{\tau_2}, \cdots d{\tau_K}) \rightarrow s_{lp}$: Returns shares $s_{lp}$ to the minted units of tokens and adds $d\tau_1, d\tau_2, \cdots d\tau_K$ tokens to liquidity pool $R\tau_1, R\tau_2, \cdots, R\tau_K$.
    \item $\func{Remove}(s_{lp}) \rightarrow (d{\tau_1}, d{\tau_2}, \cdots d{\tau_K})$: Returns $(d\tau_1, d\tau_2, \cdots d\tau_K)$ minted tokens and burns LP shares $s_{lp}$.
    \item $\func{Swap}(d\tau_i, \tau_j) \rightarrow d\tau_j$: Returns $d\tau_j$ for $d\tau_i$.
\end{itemize}
LPs can add and remove tokens to the liquidity pool by either using $\func{Add}(d\tau_1, d\tau_2, \cdots d\tau_K)$ and $\func{Remove}(s_{lp})$ (see Section~\ref{sec:add_remove}).
Traders can swap tokens using $\func{Swap}(d\tau_i, \tau_j)$ (see Section~\ref{sec:swap}).\\

We define metrics that track the states of LPs and liquidity pools. 

\begin{definition}[Total Investment Balance a.k.a, Target Balance] \label{def:tb}
    Let $A_{\ttime{1}}, A_{\ttime{2}}, \cdots, A_{\ttime{T}}$ be the sequence of actions.
    The total investment balance at the time $\ttime{T}$,
    \begin{align*}
        TB_{\ttime{T}} := \sum^T_{\ttime{t}=1} \left[ \Big(\mathbb{I}[A_{\ttime{t}}=\func{Add}] - \mathbb{I}[A_{\ttime{t}}=\func{Remove}]\Big) \sum^K_{i=1} d\tau_{i,\ttime{t}} \cdot f\tau_{i,\ttime{t}} \right]
    \end{align*}
    where $f\tau_{i,\ttime{t}}$ is the fair price and $R\tau_{i,\ttime{t}}$ is the pool balance of $\tau_i$ at time $\ttime{t}$.
\end{definition}
The total investment balance measures the amount of value that was added to the UAMM in terms of the base currency.
Because LPs can provide liquidity in $K$-minted tokens, it converts each minted token in terms of base currency and then aggregates the net amount.
LPs' total investment amount is independent of $\func{Swap}(d\tau_i, \tau_j)$ transactions.

While LPs' total investment amount does not change unless more funds are added or removed from the liquidity pool, 
the total liquidity pool value can fluctuate based on the change in fair price or liquidity pool balances.
\begin{definition}[Liquidity Pool Value] \label{def:tv}
    The total value of the liquidity pool is 
    \begin{align*}
        TV := \sum^K_{i=1} f\tau_{i} \cdot R\tau_{i}
    \end{align*}
    where $f\tau_i$ is the fair price and $R_i$ is the pool balance of $\tau_i$.
    If the fair prices are probability\footnote{the total fair prices sum up to 1 and each fair price lies in between $[0,1]$}, then the total value is the expected value (EV) of the liquidity pool.
\end{definition}

These definitions enable us to talk about the liquidity provider's impermanent gain and loss.
The impermanent gain and loss reflect the relative performance of assets within a pool compared to holding them individually over a specific time frame.
\begin{definition}[Impermanent Gain \& Loss] \label{def:tv}
    At time $T$m the impermanent gain \& loss is 
    \begin{align*}
        IGnL_{\ttime{T}}(LP) := TV_{\ttime{T}}(LP) - TB_{\ttime{T}}(LP) 
    \end{align*}
    where $TV_{\ttime{T}}(LP)$ and $TB_{\ttime{T}}(LP)$ are the liquidity pool value and total investment balance of an individual liquidity provider (LP).
    We call it gain if $IGnL_{\ttime{T}}(LP) > 0$ and loss if $IGnL_{\ttime{T}}(LP) < 0$.
\end{definition}
Impermanent gain and loss refer to situations where you deposit assets into a liquidity pool and experience a gain or loss if you were to withdraw them at a later time compared to holding those assets individually throughout the same period. 
These gains and losses become permanent only when you refund your LP shares. 

We list common properties of AMM that do not depend on the design of UAMM.
\begin{property}[Basic AMM Properties]
    Given a state $\Gamma$ and transaction $A$, 
    \begin{enumerate}[label=(\alph*)]
        \item Initial condition: At time $\ttime{0}$, $R\tau_{i,\ttime{0}} = 0$ for $\tau_i$ and $A_{\ttime{0}}=\func{Add}$.
        \item Non depletion: For $\ttime{t} >0$, $R\tau_{i,\ttime{t}} > 0$ for $\tau_i \in \mathbb{T}$ and $TV_{\ttime{t}} > 0$.
        \item Determinism: If $\Gamma \rightarrow \Gamma^\prime$ and $\Gamma \rightarrow \Gamma^{\prime\prime}$, then $\Gamma^\prime=\Gamma^{\prime\prime}$.
        \item Preservation of token supply: If $\Gamma \rightarrow \Gamma^\prime$, then total supply of tokens at $\Gamma$ and $\Gamma^\prime$ remains the same for all $\tau_i$.
    \end{enumerate}
\end{property}
The non-depletion condition enforces the AMM to never run out of liquidity and its value is greater than zero.  
The initial condition exists for the non-depletion condition where someone has to add the initial fund.
The deterministic transactions are desired properties for users and blockchains. 
It ensures that users are not taking chances and nodes can reconstruct the network and verify the validity of a transaction from a sequence of transactions.
These together set up a framework for AMMs to have a certain desirable dynamical system and allow us to do basic sanity checks.

We would like to remind you that the liquidity pool balances do not have to be preserved.
Relaxing this property makes UAMM superior to other AMMs in terms of reducing impermanent loss.
We argue that it is a direct consequence of our pricing mechanism and it is more fair to LPs in some sense (see the details in Section~\ref{sec:swap}).

%% file: add_refund.tex
\section{Adding \& Removing Tokens}
\label{sec:add_remove}

Before we dive into adding and removing funds, let us form a relationship between $(d\tau_1,d\tau_2,\cdots, d\tau_K)$ and $d\baseC{\tau_0}$.
The total value of $K$ minted tokens is $d\baseC{\tau_0}$ which is $(d\tau_1,d\tau_2,\cdots, d\tau_K)$ weighted by fair price, 
\begin{align}
    d{\baseC{\tau_0}} = \sum^{K}_{i=1} d\tau_i f\tau_i.
    \label{eq:dtau_to_dtau_i}
\end{align}
Conversely, $d\baseC{\tau_0}$ can be transformed into $(d\tau_1,d\tau_2,\cdots, d\tau_K)$. 
There are two ways to do this:
\begin{enumerate}
    \item 
A simple method is to equally divide $d\baseC{\tau_0}$ for each atomic token $\tau_i$ based on an aggregated unit of fair prices,
\begin{align}
    d\tau_i = \frac{d\baseC{\tau_0}}{\sum^{K}_{k} f\tau_k}
    \label{eq:dtau_i_to_dtau}
\end{align}
for all $\tau_i$. 
Now, $(d\tau_1,d\tau_2,\cdots, d\tau_K)$ and $d\baseC{\tau_0}$ are reversible:\\
$
    \sum^K_{i=1} d\tau_i f\tau_i = \sum^K_{i=1} \left( \frac{d\tau} {\sum^{K}_{k=1} f\tau_k} \right) f\tau_i = d\tau \sum^K_{i=1} \frac{f\tau_i} {\sum^{K}_{k=1} f\tau_k} = d\baseC{\tau_0}. 
$
\item Another way to distribute with respect to the ratio of the pool balances and the total liquidity value,
\begin{align}
    d\tau_i = d{\baseC {\tau_0} }\frac{R\tau_i}{TV}
\end{align}
for all $\tau_i$. Again, we verify that they are reversible:\\
$
    \sum^{K}_{i=1}d\tau_i \cdot f\tau_i = \sum^{K}_{i=1} d{\baseC{\tau_0}}  \cdot \frac{R\tau_i}{TV} \cdot f\tau_i 
        = \frac{d{\baseC{\tau_0}}}{TV} \sum^{K}_{i=1} R\tau_i \cdot f\tau_i = d{\baseC{\tau_0}}.
$
\end{enumerate}

\subsection*{Add \& Remove Functions}

Anyone can participate in a liquidity provision program to earn yields.
When a user adds liquidity to the pool, you receive LP tokens that represent your ownership in the liquidity pool.
Let $s_{lp}$ be the number of LP tokens that you receive when you add funds either in base currency $\baseC{\tau_0}$ or $K$ minted tokens.
$\func{Add}(d{\tau_1}, d{\tau_2}, \cdots d{\tau_K}) \rightarrow s_{lp}$ returns a new-minted amount of LP shares,
\begin{align*}
    s_{lp}= d{\baseC{\tau_0}} \cdot \frac{TS}{TV}.
\end{align*}
We update the liquidity pool balance $R\tau_i = R\tau_i + d\tau_i \text{ for all } \tau_i$, $TV$, and $TB$.

LPs can withdraw their funding from the liquidity pool anytime. 
We calculate how much $K$ minted tokens to take out and convert back to the base currency based on the number of shares $s_{lp}$.
$\func{Remove}(s_{lp}) \rightarrow (d{\tau_1}, d{\tau_2}, \cdots d{\tau_K})$ takes
$s_{lp}$ as input and returns $K$-minted tokens, where 
\begin{align*}
    d\tau_i =  R\tau_i \cdot \frac{s_{lp}}{TS}, \qquad \text{ for all } \tau_i.
\end{align*}

After performing the $\func{Add}$ or $\func{Remove}$ transaction, UAMM states will be automatically updated. 
The state $\Gamma$ consists of variables in Section~\ref{sec:state}.

\begin{property}[Additivity]
    The $\func{Add}$ and $\func{Remove}$ transactions are addictive while the fair prices remain the same.
    That is, the result of $\mathcal{A}$ and the states are the same whether a user performs two of the same successive transactions $A0$ and $A1$, or through a single transaction $A2$:
    \begin{enumerate}[label=(\alph*)]
        \item 
            $\func{Add}(d{\tau_1}, d{\tau_2}, \cdots d{\tau_K}) + \func{Add}(d{\tau_1}^\prime, d{\tau_2}^\prime, \cdots d{\tau_K}^\prime) 
                    \Longleftrightarrow \func{Add}(d{\tau_1}+d{\tau_1}^\prime, d{\tau_2}+d{\tau_2}^\prime, \cdots , d{\tau_K}+d{\tau_K}^\prime)$
        \item $\func{Remove}(s_{lp}) + \func{Remove}(s_{lp}^\prime) \Longleftrightarrow \func{Remove}(s_{lp}+s_{lp}^\prime)$
        \item $\Gamma \xrightarrow{A0} \Gamma_0 \xrightarrow{A1} \Gamma_1 \Longleftrightarrow \Gamma \xrightarrow{A2} \Gamma_1$
        where $A0=A1=A2 \in\lbrace \func{Add}, \func{Remove} \rbrace$ 
    \end{enumerate}
\end{property}

\begin{property}[Reversibility]
    The $\func{Add}$ and $\func{Remove}$ transactions are reversible while the fair prices remain the same.
    The state derived from $\func{Add}$ operation is reversible by $\func{Remove}$, and visa versa:
    \begin{enumerate}[label=(\alph*)]
        \item $\func{Add}(\func{Remove}(s_{lp})) \rightarrow s_{lp}$
        \item $\func{Remove}(\func{Add}(d{\tau_1}, d{\tau_2}, \cdots d{\tau_K})) \rightarrow (d{\tau_1}, d{\tau_2}, \cdots d{\tau_K})$
        \item If $\Gamma \xrightarrow{A} \Gamma^\prime$, there exist $A^{-1}$ such that $\Gamma^\prime \xrightarrow{A^{-1}} \Gamma$ where
        $A, A^{-1} \in\lbrace \func{Add}, \func{Remove} \rbrace$.
    \end{enumerate}
\end{property}
The derivations of additivity and reversibility are shown in Appendix~\ref{app:add_remove}.
The liquidity pool balances are addictive and reversible regardless of the fair prices.
However, the liquidity value (TV) and total invested balance (TB) differ when fair prices change.
Both additivity and reversibility ensure that LPs cannot take advantage or disadvantage of add and remove operations within the same fair prices.

%% file: swap_function.tex
\section{Swap Transaction}
\label{sec:swap}

The core idea of UAMM is fundamentally different from the rest of other AMMs.
AMMs like Uniswap compute the price of $\tau_i$ for $\tau_j$ based on liquidity pool balances.
Because it only considers internal market data, the Uniswap price can be far off from the prices in other exchanges.
Especially the internal market is inefficient when the internal market volume is relatively small to the external market.
This creates arbitrage opportunities for traders and potential losses for LPs.
In contrast, we respect efficient market theory and estimate fair prices given the external and internal market data. 
We use the UAMM algorithm to calculate the slippage while referencing fair price as a spontaneous price.
In this section. we describe when and how to add slippage such that we prevent impermanent loss for LPs.

\subsection*{Notations}
\noindent A user can swap $d\tau_i$ tokens for some units of $\tau_j$ through UAMM. 
Because swapping between $\tau_i$ and $\tau_j$ applies for any $0\leq i,j \leq K$, 
without loss of generality, we label $\tau_{\text{In}}$ and $\tau_{\text{Out}}$ for input and output token index.

\begin{itemize}
    \item Let $R\tau_{\text{In}}$ and $R\tau_{\text{Out}}$ be the liquidity pool balance of input and output tokens.
    \item Let $f\tau_{\text{In}}$ and $f\tau_{\text{Out}}$ be the probability of input and output event.
    \item Let $d\tau_{\text{In}}$ be the input token amount before adding slippage.
    \item Let $\Delta\tau_{\text{Out}}$ be the output token amount before adding slippage.
    \item Let $d\tau_{\text{Out}} = g(\Delta\tau_{\text{Out}})$ be the output token amount with slippage.
\end{itemize}

\subsection*{Swapping input token for output token}

Given the fair prices $f\tau_{\text{In}}$ and $f\tau_{\text{Out}}$, the fair exchange without any slippage would be
\begin{align}
    d\tau_{\text{Out}} = \rho \cdot d\tau_{\text{In}}
    \label{eq:fair_exchange}
\end{align}
where $\rho = \frac{f\tau_\text{In}}{f\tau_\text{Out}}$ is the fair swapping rate.
Unfortunately, there is a danger of liquidity pool balances drying up if the demand for the $\tau_{\text{Out}}$ heavily outweighs the demand for the $\tau_{\text{In}}$.
For this reason, we add slippage to the swapping rate such that it respects the liquidity pool balance and the user's trade size.

Now, we describe the intuition behind how we derive the UAMM swapping function. 
The amount of output token $d\tau_{\text{Out}}$ depends on the input token $d\tau_{\text{In}}$, the liquidity pool situation $R\tau_{\text{In}}$ and $R\tau_{\text{Out}}$, and
the target balance $TB$ where $TB$ is the desired balance for all liquidity pool balances.
The price calculation has the following form,
\begin{align*}
    d\tau_{\text{Out}} &=  \underbrace{\func{Swap} \Big( \underbrace{ \rho \cdot  d\tau_{\text{In}}}_{\Delta\tau_{\text{Out}}} ; \Gamma \Big)}_{\text{Output Slippage }}
\end{align*}
where $\Gamma$ is the state parameters and $\rho = \frac{\rho\tau_{\text{In}}}{\rho\tau_{\text{Out}}}$ is the price ratio.
Intuitively, we want to add slippage to output pools $R\tau_{\text{Out}}$ if we are below the desired target balance $TB$. 
This means that we compute the slippage for each pool, starting with the input liquidity pool $R\tau_{\text{In}}$ and then the output liquidity pool $R\tau_{\text{Out}}$. Here is the following steps that to compute $d\tau_{\text{Out}}$:
\begin{enumerate}
    \item compute the swapping amount based on price ratio, $\Delta\tau_{\text{Out}} = \rho \cdot d\tau_{\text{In}}$. 
    \item compute the slippage on output pool $R\tau_{\text{Out}}$, $d\tau_{\text{Out}} = \func{swap}(\Delta\tau_{\text{Out}}; \Gamma)$.
\end{enumerate}
Our liquidity pool balance is in a good state when our pool balance is above the target balance
and the slippage is only applied when our pool balance $R\tau_i$ is below our target balance $TB$.
Thus, the slippage function $g(d\tau_{\text{Out}}; \Gamma)$ is a piece-wise function such that
\begin{align*}
    \func{swap}(\Delta\tau_{\text{Out}}; \Gamma) =
    \begin{cases} 
        \alpha \cdot \Delta\tau_\text{Out} + (\rho -\alpha) (R\tau_\text{Out}-TB) & \text{ if } R\tau_{\text{Out}} - \Delta\tau_\text{Out}\leq TB \leq R\tau_{\text{Out}} \\
        \Delta\tau_{\text{Out}} & \text{ else if } TB \leq R\tau_{\text{Out}} \\
        R\tau_{\text{Out}} - \frac{TB^2}{X\tau_{\text{Out}}+\Delta\tau_\text{Out}}  & \text{ otherwise }
    \end{cases}
\end{align*}
where $\alpha= \frac{R\tau_{\text{Out}}}{X\tau_{\text{Out}}+ \Delta\tau_{\text{Out}}}$ and $TB^2 = X\tau_{\text{Out}}\cdot R\tau_{\text{Out}}$ is a target balance of the pool.
The first if statements in $\func{swap}$ are to ensure that the functions are continuous with respect to $d\tau_{\text{In}}$ and $\Delta\tau_{\text{Out}}$ respectively.

The above slippage function is derived from the constant product curve with constant $TB^2$ when the output liquidity pool $R\tau_\text{Out}$ is below the 
target balance $TB$. If $TB > R\tau_{\text{Out}}$, our constant product curve for the output slippage function $g(\Delta_{\text{Out}}; \Gamma)$ is 
\begin{align*}
    TB^2 
        &= \left( R\tau_{\text{Out}} - \frac{R\tau_{\text{Out}}}{X\tau_{\text{Out}}+\rho d\tau_{\text{In}}} \right) \left( X\tau_{\text{Out}} + \rho d\tau_{\text{In}} \right)
\end{align*}
Technically, you can also add the slippage to the input function in a similar manner when the input liquidity pool $R\tau_\text{In} > TB$. 
The hope is that it discourages users from executing the trade and pulls the $R\tau_\text{In}$ towards $TB$.
We show the derivation for the input slippage function in Appendix~\ref{app:swap_func}.


Formally, the slippage rate incorporated in the swapping function has the following expression:
\begin{align}
    d\tau_{\text{Out}} = \rho \cdot USX(d\tau_{\text{In}}, \Gamma) \cdot d\tau_{\text{In}}
    \label{eq:uamm_swap_func}
\end{align}
The only difference between the fair swap function in Equation~\ref{eq:fair_exchange} and the UBET swap function in Equation~\ref{eq:uamm_swap_func} is the UBET slippage rate.
$USX$ is a UAMM slippage rate defined below.
\begin{definition}[UBET slippage rate]
    The UBET Automated Market Maker rate function is
    \begin{align}
        USX(d\tau_{\text{In}}, \Gamma) = 
        \begin{cases}
            1 & \text{ if $TB < R\tau_{\text{Out}}$}\\
            \xi & \text{ if }R\tau_{\text{Out}} < TB \leq R\tau_{\text{Out}} + d\tau_\text{In}\\
            \frac{R\tau_{\text{Out}}}{X\tau_{\text{Out}}+\rho d\tau_{\text{In}}} & \text{ if }  TB \geq R\tau_{\text{Out}}
        \end{cases}
        \label{eq:ubet_slippage_rate}
    \end{align}
    where $\xi(d\tau_{\text{In}}) =  \left(R\tau_\text{Out}-TB\right)\rho + \left(d\tau_\text{In} - \left(R\tau_\text{Out}-TB\right)\right) \frac{R\tau_{\text{Out}}}{X\tau_{\text{Out}}+\rho d\tau_{\text{In}}}$.
\end{definition}
TODO: slippage definition.

Using the UBET slippage rate, we can assure the following properties of the swap transaction.
The output of the swapping function $d\tau_\text{Out}$ is bounded by $R\tau_\text{Out}$ and this assures the non-depletion property of AMM.
\begin{property}[Output-boundedness]
    UAMM always have enough output tokens $R\tau_\text{Out}$ for a user to perform swap $\tau_\text{In}$ for $\tau_\text{Out}$,
    \begin{align*}
        0 \leq d\tau_{\text{Out}} = \rho \cdot USX(d\tau_{\text{In}}, \Gamma) \cdot d\tau_{\text{In}} < R\tau_\text{Out}
    \end{align*}
    for all $d\tau_\text{In} \geq 0$ and $R\tau_\text{In}, R\tau_\text{Out} > 0$.
\end{property}

The monotonicity of the swap function ensures that the gain or loss of the user after performing the swap is also monotonic.
\begin{property}[Monotonicity]
UAMM slippage rate is monotonic:
\begin{enumerate}[label=(\alph*)]
    \item 
        $USX(d\tau_\text{In}, \Gamma)  \leq USX(d\tau_\text{In}^\prime, \Gamma)$ if $d\tau_\text{In} \leq d\tau_\text{In}^\prime$ and $\Gamma$ is fixed.
    \item 
        $USX(d\tau_\text{In}, \Gamma)  \leq USX(d\tau_\text{In}^\prime, \Gamma^\prime)$ if $d\tau_\text{In} \leq d\tau_\text{In}^\prime$, $R\tau_\text{In} \leq R\tau_\text{In}^\prime$, $R\tau_\text{Out} \leq R\tau_\text{Out}^\prime$ and the rest are fixed. 
\end{enumerate}
\end{property} 
UAMM swap function does not add slippage with respect to the input but only the output liquidity pool. 
This allows us to have a monotonic slippage rate.
It is possible to include the input slippage function (see Appendix~\ref{app:uamm_swap_input_usx}).
However, once you add this input slippage function, we do not get a monotonic function as the input liquidity pool increase, $R\tau_{In} < R\tau_{In}^\prime$. This is because the input slippage function pulls back the liquidity pool balances so that it does not get too distant from the target balance $TB$. 
We advise that it is not necessary to have an input slippage $\Delta\tau_\text{In}$ and it can simply be removed (i.e., $\Delta\tau_\text{In}=d\tau_\text{In}$).
This is because input slippage $\Delta\tau_\text{In}$ can be overly too strict in the sense that it charges too much slippage and it prevents having an impermanent gain. 

\begin{property}[Homogeneity]
    UAMM slippage rate $USX$ is homogeneous, for $a > 0$,
    \begin{align*}
        USX(a\cdot d\tau_\text{In}, \Gamma^\prime) = USX(d\tau_\text{In}, \Gamma).
    \end{align*}
    where $\Gamma = (R\tau_1, R\tau_2, \cdots, R\tau_K, TB)$ and $\Gamma^\prime = (a\cdot R\tau_1, a \cdot R\tau_2, \cdots, a \cdot R\tau_K, a\cdot TB)$.
\end{property}
Typically, homogeneity property leads to equal spontaneous price for $\Gamma$ and $\Gamma^\prime$ for AMMs like Uniswap \cite{Adams2020UniswapVC}, which
 maintains the ratio of the liquidity pool balances when you add or remove minted tokens. This is because their spontaneous price directly depends on the ratio of the pool.
 However, our spontaneous price is determined by the fair price, not by the liquidity pool balances, and only the slippage rate depends on the liquidity pool balances. 
 Therefore, the homogeneity of $USX$ is irrelevant to the homogeneity of spontaneous price after performing add or remove functions unless the deposit and withdrawal amount maintains the pool balance ratios.

 \begin{property}[Additivity]
    The $\func{Swap}$ transaction is additive while the fair prices remain the same.
    \begin{enumerate}[label=(\alph*)]
        \item UAMM swap function is additive
            \begin{align*}
                \func{swap}(d\tau_\text{In}; \Gamma) + \func{swap}(d\tau_\text{In}; \Gamma^\prime) 
                    = \func{swap}(d\tau_\text{In}+d\tau_\text{In}^\prime; \Gamma)
            \end{align*}    
            where 
            $\Gamma \xrightarrow{\func{swap}(d\tau_\text{In})} \Gamma^\prime \xrightarrow{\func{swap}(d\tau_\text{In}^\prime)} \Gamma^{\prime\prime}$ are two output slippage functions. 
        \item 
            Let $\alpha = USX(d\tau_\text{In}, \tau_\text{Out}; \Gamma)$
            and let $\beta = USX(d\tau_\text{In}^\prime, \tau_\text{Out}; \Gamma^\prime)$. Then,
            \begin{align*}
                    USX(d\tau_\text{In}+d\tau_\text{In}^\prime, \tau_\text{Out}) = \frac{\alpha \cdot d\tau_\text{In}+ \beta \cdot d\tau_\text{In}^\prime}{d\tau_\text{In}+d\tau_\text{In}^\prime}
            \end{align*}    
        \item The states are the same whether a user performs two of the same successive swap transactions, or through a single swap transaction:
                $\Gamma \xrightarrow{\func{swap}(d\tau_\text{In}+d\tau_\text{In}^\prime)} \Gamma_1$.
    \end{enumerate}
\end{property}
The proof is shown in Appendix~\ref{proof:addivitiy_thm}.
This property applies to multiple sequences of swap transactions.
The amount of slippage and a user's gain or loss will be the same whether you make single or multiple consecutive transactions.

As a side note, having an input slippage function as shown in Appendix~\ref{app:uamm_swap_input_usx} does not guarantee additivity.
Indeed, having a single swap transaction with $d\tau_\text{In}+d\tau_\text{In}^\prime$ leads to the difference of $d\tau_\text{In} \cdot d\tau_\text{In}^\prime$ compare to having two successive transactions. 

In general, UAMM $swap$ transaction is irreversible.
However, we still assure you that the $swap \circ swap$ transaction will end up with less amount than the originally invested amount.
\begin{property}[Weak-Reversibility]
    $\func{swap} \circ \func{swap}(d\tau_\text{In}; \Gamma) \leq d\tau_\text{In}$ 
\end{property}
This is a desirable property, otherwise, the user can take arbitrage out of the liquidity funds.
The user should not be able to get more than their original amount from reverse transactions.
Moreover, UAMM is reversible when our liquidity pool balances are in a good situation where
$R\tau_\text{Out} -\Delta\tau_\text{Out} > TB$ and $R\tau_\text{In} > TB$. 
This situation indicates that LPs already have made a good profit out of previous transactions.

%% file: spontaneous_price.tex
\section{Spontaneous Price}
The spontaneous price is our market price with no slippage. 
One  can consider the spontaneous price as a fair price with an applied spread.
Because the price depends on whether $R\tau_{\text{In}}$ and $R\tau_{\text{Out}}$ are below the target price or not,
the spontaneous price is also a piece-wise function, 
\begin{align}
    \frac{p\tau_\text{In}}{p\tau_\text{Out}} = \left(\frac{\text{d} d\tau_{\text{In}}} {\text{d} d\tau_{\text{Out}}}\Bigg|_{d\tau_{\text{Out}}=0}\right)^{-1}
\end{align}
where
\begin{align*}
    \frac{\text{d} d\tau_{\text{In}}} {\text{d} d\tau_{\text{Out}}}\Bigg|_{d\tau_{\text{Out}}=0} =
        \begin{cases}
            \frac{1}{\rho} & \text{ if $TB \leq R\tau_{\text{In}}$ and $TB \leq R\tau_{\text{Out}}$  }\\
            \frac{X\tau_{\text{Out}}}{\rho \cdot R\tau_{\text{Out}}} & \text{ if }  TB > R\tau_{\text{Out}} \text{ and } TB \leq R\tau_{\text{In}}\\
        \end{cases}
\end{align*}
The subscript $|_{d\tau_{\text{Out}}=0}$ illustrates what happens to the price of an infinitesimally small trade.
The derivation is shown in the Appendix.
It is also easy to see that the UAMM slippage rate together with $\rho$ converges to the spontaneous price,
\begin{property}[Convergence]
$\frac{p\tau_\text{In}}{p\tau_\text{Out}} = \lim_{d\tau_\text{In}\rightarrow 0} \rho \cdot USX(d\tau_\text{In}, \Gamma)$.
\end{property}

\begin{property}
    UAMM swap rate is always less than or equal to the spontaneous price rate and the spontaneous price rate is lower than or equal to the fair swap rate,
    \begin{align*}
        \rho \cdot USX(d\tau_\text{In}, \Gamma) \leq \frac{p\tau_\text{In}}{p\tau_\text{Out}} \leq \frac{f\tau_\text{In}}{f\tau_\text{Out}} = \rho
    \end{align*}
    for all $d\tau_\text{In} > 0$.
\end{property}

%% file: conclusion.tex
\section{Conclusion}

In conclusion, the existing AMMs used in DEX suffer from a significant drawback—arbitrage opportunities. This occurs due to the AMMs' reliance on internal market information, disregarding external market prices and leading to potential losses for LPs. Additionally, AMMs have slower adaptation to price changes, relying on traders to exploit arbitrage opportunities for price adjustments.

To address these issues, the UAMM algorithm is proposed. UAMM calculates prices by considering both external and internal market prices, effectively eliminating arbitrage opportunities. By minimizing impermanent loss and maintaining the desired properties of a constant product curve, UAMM provides LPs with improved risk management and passive yield generation.

With UAMM, LPs can trade with increased security and stability, as they no longer need to worry about impermanent loss management. The algorithm defines three transactions, add, remove, swap, and their properties induce desired behaviours for the AMM. Furthermore, UAMM exhibits zero arbitrage opportunities when the market price aligns with the fair price assumption, further enhancing LPs' confidence and profitability.

The UAMM algorithm presents a promising solution to the limitations of existing AMMs in decentralized exchanges. By considering external market prices and implementing measures to minimize impermanent loss, UAMM enhances the efficiency and reliability of DEX while providing LPs with a more secure and profitable trading experience.

%% file: appendix.tex
\appendix

\section{Appendix}
\setcounter{property}{1}

\subsection{Adding \& Removing Funds}\label{app:add_remove}
\begin{property}[Additivity]
    The $Add$ and $Remove$ transactions are addictive while the fair prices remain the same.
    That is, the result of $\mathcal{A}$ and the states are the same whether a user performs two of the same successive transactions $A0$ and $A1$, or through a single transaction $A2$:
    \begin{enumerate}
        \item 
            $Add(d{\tau_1}, d{\tau_2}, \cdots d{\tau_K}) + Add(d{\tau_1}^\prime, d{\tau_2}^\prime, \cdots d{\tau_K}^\prime) 
                    \Longleftrightarrow Add(d{\tau_1}+d{\tau_1}^\prime, d{\tau_2}+d{\tau_2}^\prime, \cdots , d{\tau_K}+d{\tau_K}^\prime)$
        \item $Remove(s_{lp}) + Remove(s_{lp}^\prime) \Longleftrightarrow Remove(s_{lp}+s_{lp}^\prime)$
        \item $\Gamma \xrightarrow{A0} \Gamma_0 \xrightarrow{A1} \Gamma_1 \Longleftrightarrow \Gamma \xrightarrow{A2} \Gamma_1$
        where $A0=A1=A2 \in \mathcal{A}\in\lbrace Add, Remove \rbrace$ 
    \end{enumerate}
\end{property}
\begin{proof}
\hfill 
\begin{enumerate}
    \item 
    Let $s_{lp} \leftarrow Add(d{\tau_1}, d{\tau_2}, \cdots d{\tau_K})$.\\
    Let $s_{lp}^\prime \leftarrow Add(d{\tau_1}^\prime, d{\tau_2}^\prime, \cdots d{\tau_K}^\prime)$.\\
    Let $s_{lp}^{\prime\prime} \leftarrow Add(d{\tau_1}+d{\tau_1}^\prime, d{\tau_2}+d{\tau_2}^\prime, \cdots , d{\tau_K}+d{\tau_K}^\prime)$.\\
    Suppose two of the same successive transactions, $Add(d{\tau_1}, d{\tau_2}, \cdots d{\tau_K})$ and $Add(d{\tau_1}^\prime, d{\tau_2}^\prime, \cdots d{\tau_K}^\prime)$.
    \begin{align*}
        s_{lp}+s_{lp}^\prime &= d\tau_0\frac{TS}{TV} + d\tau_0^\prime\frac{TS+s_{lp}}{TV+d\tau_0}\\
                    &= \frac{d\tau_0\cdot TS (TV+d\tau_0) + d\tau_0^\prime \cdot TV (TS+s_{lp})}{TV(TV+d\tau_0)}\\
                    &= \frac{d\tau_0\cdot TS (TV+d\tau_0) + d\tau_0^\prime \cdot TV (TS+d\tau_0 \frac{TS}{TV})}{TV(TV+d\tau_0)}\\
                    &= \frac{TS}{TV} \frac{ d\tau_0 (TV+d\tau_0) + d\tau_0^\prime (TV+d\tau_0) }{TV+d\tau_0}\\
                    &= \frac{TS}{TV}  \left(d\tau_0 + d\tau_0^\prime \right)\\
                    &= s_{lp}^{\prime\prime}
    \end{align*}
    In the first equality, we used the definition of the $Add$ function and the fact that $d\tau_0 = \sum^K_{k=1} d\tau_k f\tau_k$. 
    \item 
    Let $(d{\tau_1}, d{\tau_2}, \cdots d{\tau_K}) = Remove(s_{lp})$.\\
    Let $(d{\tau_1}^\prime, d{\tau_2}^\prime, \cdots d{\tau_K}^\prime) = Remove(s_{lp}^\prime)$.\\
    Let $(d{\tau_1}^{\prime\prime}, d{\tau_2}^{\prime\prime}, \cdots , d{\tau_K}^{\prime\prime}) = Remove(s_{lp}^{\prime\prime})$.
    \begin{align*}
        d\tau_i + d\tau_i^\prime &= R\tau_i \frac{s_{lp}}{TS} + (R\tau_i - d\tau_i) \frac{s_{lp}^\prime}{TS-s_{lp}}\\
                        &= \frac{R\tau_i \cdot s_{lp} \cdot (TS-s_{lp}) + (R\tau_i-d\tau_i)\cdot s_{lp}^\prime \cdot TS}{TS(TS-s_{lp})}\\
                        &= \frac{R\tau_i \cdot s_{lp} \cdot (TS-s_{lp}) + (R\tau_i-d\tau_i)\cdot R\tau_i \frac{s_{lp}}{TS} \cdot TS}{TS(TS-s_{lp})}\\
                        &= \frac{R\tau_i}{TS} \frac{s_{lp} \cdot (TS-s_{lp}) + s_{lp}^\prime \cdot (TS-s_{lp})}{TS-s_{lp}}\\
                        &= \frac{R\tau_i}{TS} \left( s_{lp} + s_{lp}^\prime \right)  \\
                        &= d\tau_i^{\prime\prime}
    \end{align*}
    \item 
    The liquidity pool balances are addictive and reversible regardless of the fair prices (trivial).
    We verify that the total investment balance and liquidity pool value at $T$ remain the same.
    Let $A0$, $A1$, and $A2$ be the add function.
    Without loss of generality, $A0$ and $A1$ occurred on time step $T-2$ and $T-1$, then 
    \begin{align*}
        TB_{T-2}\xrightarrow{A0,A1} TB^\prime &= \sum^T_{t=1} \left[ \Big(\mathbb{I}[A_t=Add] - \mathbb{I}[A_t=Remove]\Big) \sum^K_{i=1} d\tau_{i,t} f_{\tau_{i,t}} \right]\\
             &=  TB_{T-2} + \left[\mathbb{I}[A_{T-2}=A0] \sum^K_{i=1} d\tau_{i,T-2} f\tau_{i,T-2} \right] + \left[\mathbb{I}[A_{T-1}=A1] \sum^K_{i=1} d\tau_{i,T-1}^\prime f\tau_{i,T-1} \right]\\
             &= TB_{T-2} + \sum^K_{i=1} d\tau_{i} f\tau_{i,T-2} + d\tau_{i}^\prime f\tau_{i,T-1} \\
             &= TB_{T-2} + \sum^K_{i=1} (d\tau_{i}  + d\tau_{i}^\prime) f\tau_{i,T-2} \\
             &= TB_{T-2} + \mathbb{I}[A_{T-2}=A2] \sum^K_{i=1} (d\tau_{i}  + d\tau_{i}^\prime) f\tau_{i,T-2} \\
             &= TB_{T-2}\xrightarrow{A2} TB^\prime
    \end{align*}
    The second last equality is due to fair prices not changing $f\tau_{i,T-2}= f\tau_{i,T-1}$.
    The last equality shows that it is equivalent to doing a single transaction $A_2$ at time step $T-2$.

    Let's look at the total investment amount under a similar situation.
    The total value after $A0$ and $A1$ would be
    \begin{align*}
    TV \xrightarrow{A0,A1} TV\prime &= \sum^K_{i=1} f\tau_{i,T-2} R\tau_i + f\tau_{i,T-2} d\tau_i + f\tau_{i,T-1} d\tau_i^\prime\\
       &= \sum^K_{i=1} f\tau_{i,T-2} R\tau_i + f\tau_{i,T-2} d\tau_i + f\tau_{i,T-2} d\tau_i^\prime\\
       &= \sum^K_{i=1} f\tau_{i,T-2} (R\tau_i + d\tau_i + d\tau_i^\prime)\\
       &= TV \xrightarrow{A2} TV^\prime.
    \end{align*}
    The last equality is the total liquidity pool value after applying $A2$ transaction.
\end{enumerate}
\end{proof}

\begin{property}[Reversibility]
    The $Add$ and $Remove$ transactions are reversible while the fair prices remain the same.
    The state derived from $Add$ operation is reversible by $Remove$, and visa versa:
    \begin{enumerate}
        \item $Add(Remove(s_{lp})) \rightarrow s_{lp}$
        \item $Remove(Add(d{\tau_1}, d{\tau_2}, \cdots d{\tau_K})) \rightarrow (d{\tau_1}, d{\tau_2}, \cdots d{\tau_K})$
        \item If $\Gamma \xrightarrow{A} \Gamma^\prime$, there exist $A^{-1}$ such that $\Gamma^\prime \xrightarrow{A^{-1}} \Gamma$ where
        $A\in\lbrace Add, Remove \rbrace$.
    \end{enumerate}
\end{property}
\begin{proof}
    \hfill
    \begin{enumerate}
        \item 
        Let $s_{lp} = Add(d{\tau_1}, d{\tau_2}, \cdots d{\tau_K})$.
        We apply $Remove(s_{lp})$:
        \begin{align*}
            Remove(s_{lp}) &= (R\tau_i+d\tau_i) \cdot \frac{s_{lp}}{TS+s_{lp}}\\
                &= (R\tau_i+d\tau_i) \cdot \frac{d\tau_0 \frac{TS}{TV}}{TS+d\tau_0 \frac{TS}{TV}}\\
                &= (R\tau_i+d\tau_i) \cdot \frac{d\tau_0 \frac{TS}{TV}}{TS+d\tau_0 \frac{TS}{TV}}\\
                &= d\tau_0 \cdot \frac{(R\tau_i+d\tau_i)}{(TV+d\tau_0)}\\
                &= d\tau_i^\prime
        \end{align*}
        for all $i=1,\cdots,K$.
        To verify that $(d\tau_1^\prime, d\tau_2^\prime, \cdots, d\tau_K^\prime)$ is worth the same as $(d{\tau_1}, d{\tau_2}, \cdots d{\tau_K})$,
        we first show that $d\tau_0 = \sum^{K}_{i=1} d\tau_i^\prime \cdot f\tau_i = \sum^{K}_{i=1} d\tau_i \cdot f\tau_i$:
        \begin{align*}
            \sum^{K}_{i=1} d\tau_i^\prime \cdot f\tau_i &= \sum^{K}_{i=1}  \frac{(R\tau_i+d\tau_i)}{(TV+d\tau_0)} \cdot d\tau_0 \cdot f\tau_i 
                            = \frac{d\tau_0}{(TV+d\tau_0)} \sum^{K}_{i=1} (R\tau_i+d\tau_i) \cdot f\tau_i
                            = d\tau_0
        \end{align*}
        Since the fair prices $f\tau_i$ didn't change and the values $d\tau_0$ are the same, $d\tau_i=d\tau_i^\prime$ for all $i=1,\cdots, K$.
        \item 
        Let $(d{\tau_1}, d{\tau_2}, \cdots d{\tau_K}) = Remove(s_{lp})$. 
        We apply $Add(d{\tau_1}, d{\tau_2}, \cdots d{\tau_K})$:
        \begin{align*}
            Add(d{\tau_1}, d{\tau_2}, \cdots d{\tau_K}) &= \frac{TS}{TV}\sum^{K}_{i=1} d\tau_i \cdot f\tau_i\\
                         &= \frac{TS}{TV}\sum^{K}_{i=1} \frac{s_{lp}}{TS} R\tau_i \cdot f\tau_i\\
                         &= s_{lp} \frac{TS}{TV} \frac{1}{TS} \sum^{K}_{i=1} R\tau_i \cdot f\tau_i\\
                         &= s_{lp}.
        \end{align*}
        The last equality is using the fact that $TV=\sum^{K}_{i=1} R\tau_i \cdot f\tau_i$.
        \item Using (1) and (2), it is trivial to show that there exists an inverse function that always reverses the origin state for TB and TV.
    \end{enumerate}
\end{proof}

\subsection{Swapping input token for output token}
\label{app:swap_func}

\begin{property}[Output-boundedness]
    UAMM always have enough output tokens $R\tau_\text{Out}$ for a user to perform swap $\tau_\text{In}$ for $\tau_\text{Out}$,
    \begin{align*}
        0 \leq d\tau_{\text{Out}} = \rho \cdot USX(d\tau_{\text{In}}, \Gamma) \cdot d\tau_{\text{In}} < R\tau_\text{Out}
    \end{align*}
    for all $d\tau_\text{In} \geq 0$ and $R\tau_\text{In}, R\tau_\text{Out} > 0$.
\end{property}
\begin{proof}
    It is easy to see that $d\tau_\text{Out}$ is less than $R\tau_\text{Out}$ when the trade size $\rho d\tau_{In} < R\tau_\text{Out}$.
    Consider case $\rho d\tau_\text{In} > R\tau_\text{Out}$.  
    The argument is that even if a fair exchange rate with input slippage is greater than the liquidity in the output pool, then output slippage ensures that $d\tau_\text{Out}$ is still less than the output liquidity pool $R\tau_\text{Out}$.
    If $TB \geq R\tau_\text{Out}$, then
    \begin{align*}
        d\tau_\text{Out} = \rho \cdot USX(d\tau_{\text{In}}, \Gamma) d\tau_\text{In} & = \rho \cdot \left( \frac{R\tau_\text{Out}}{X\tau_\text{Out}+\rho d\tau_\text{In}}  \right) \cdot d\tau_\text{In} \\
                &= R\tau_\text{Out} \frac{\rho \cdot d\tau_\text{In}}{X\tau_\text{Out}+\rho d\tau_\text{In}}\\
                    &= R\tau_\text{Out} \frac{1}{1+\frac{X\tau_\text{Out}}{\rho \cdot d\tau_\text{In}}}\\
                    &< R\tau_\text{Out}
    \end{align*}
    If we put $d\tau_\text{In}$ to the output slippage function, then similar to above, we will add output slippage.

    We can do better by giving $\frac{1}{\rho}(R\tau_\text{Out} - TB)$ at a fair exchange rate and the remaining $(\Delta\tau_\text{Out} - (R\tau_\text{Out}-TB)$ is traded with output slippage,
        \begin{align*}
        d\tau_\text{Out} &= \rho \cdot USX(TB, \Gamma) d\tau_\text{In}\\ 
                &= (R\tau_\text{Out} - TB) + \rho \cdot \left( \frac{R\tau_\text{Out}}{X\tau_\text{Out}+\rho TB} \right) \cdot TB \\
                &= (R\tau_\text{Out} - TB) +R\tau_\text{Out} \frac{\rho \cdot TB}{X\tau_\text{Out}+\rho TB}\\
                    &=  (R\tau_\text{Out} - TB) + TB  \underbrace{\frac{1}{1+\frac{X\tau_\text{Out}}{\rho \dot TB}}}_{<1}\\
                    &<  (R\tau_\text{Out} - TB)  + TB\\
                    &=  R\tau_\text{Out}
    \end{align*}
    The second last equality comes from the fact that our output liquidity pool has $TB$ left after giving up $\frac{1}{\rho}(R\tau_\text{Out} - TB)$ at a fair exchange rate.
\end{proof}

\begin{property}[Homogeneity]
    UAMM slippage rate $USX$ is homogeneous, for $a > 0$,
    \begin{align*}
        USX(a\cdot d\tau_\text{In}, \Gamma^\prime) = USX(d\tau_\text{In}, \Gamma).
    \end{align*}
    where $\Gamma = (R\tau_\text{In}, R\tau_\text{Out}, TB)$ and $\Gamma^\prime = (a\cdot R\tau_\text{In}, a \cdot R\tau_\text{Out}, a\cdot TB)$.
\end{property}
\begin{proof}
    The derivation is simply plugging in the scalar to $USX(a\cdot d\tau_\text{In}, \Gamma^\prime)$.
    We show an example when case (*): $TB > R\tau_\text{Out}$ and $TB < R\tau_\text{In}$.
    We can extend the proof for the input slippage function that is defined in Appendix~\ref{app:uamm_swap_input_usx}.
    First, we can write $a \cdot \Delta\tau_\text{In} = f(a\cdot d\tau_\text{In}, \Gamma)$ since
    \begin{align*}
    f(a\cdot d\tau_{\text{In}}; \Gamma) =
        \begin{cases} 
            a\cdot d\tau_{\text{In}} & \text{ if } TB \geq R\tau_{\text{In}} \\
            a\cdot d\tau_{\text{In}}\frac{ a \cdot (R_{\text{In}} + d\tau_{\text{In}})}{a\cdot X\tau_{\text{In}}} & \text{ otherwise }.
        \end{cases}
    \end{align*}
    Now, we can write the $USX$ for the case (*) in terms of $a\Delta\tau_\text{In}$,
    \begin{align*}
        USX(a\cdot d\tau_\text{In}, \Gamma) &= \frac{ a\cdot \Delta\tau_{\text{In}}}{a\cdot d\tau_{\text{In}}} \left( \frac{a\cdot R\tau_{\text{Out}}}{a\cdot X\tau_{\text{Out}}+\rho a\cdot \Delta\tau_{\text{In}}} \right)\\
            &= \frac{ \Delta\tau_{\text{In}}}{d\tau_{\text{In}}} \left( \frac{R\tau_{\text{Out}}}{X\tau_{\text{Out}}+\rho \Delta\tau_{\text{In}}} \right)
            = USX(d\tau_\text{In}, \Gamma) 
    \end{align*}
    for case (*). 
    You can do the same exercise for the rest of other cases in $USX$.
\end{proof}

 \begin{property}[Additivity]
    \label{proof:addivitiy_thm}
    The $Swap$ transaction is additive while the fair prices remain the same.
    \begin{enumerate}[label=(\alph*)]
        \item UAMM swap function is additive
            \begin{align*}
                swap(\Delta\tau_\text{Out}; \Gamma) + swap(\Delta\tau_\text{Out}; \Gamma^\prime) 
                    = swap(\Delta\tau_\text{Out}+\Delta\tau_\text{Out}^\prime; \Gamma)
            \end{align*}    
            where 
            $\Gamma \xrightarrow{swap(\Delta\tau_\text{Out}, \Gamma)} \Gamma^\prime \xrightarrow{swap(\Delta\tau_\text{Out}^\prime, \Gamma^\prime)} \Gamma^{\prime\prime}$ are two output slippage functions and $\Delta\tau_\text{Out} = \rho d\tau_\text{In}$.
        \item 
            Let $\alpha = USX(d\tau_\text{In}, \tau_\text{Out}; \Gamma)$.\\
            Let $\beta = USX(d\tau_\text{In}^\prime, \tau_\text{Out}; \Gamma^\prime)$.
            \begin{align*}
                    USX(d\tau_\text{In}+d\tau_\text{In}^\prime, \tau_\text{Out}) = \frac{\alpha \cdot d\tau_\text{In}+ \beta \cdot d\tau_\text{In}^\prime}{d\tau_\text{In}+d\tau_\text{In}^\prime}
            \end{align*}    
        \item The states are the same whether a user performs two of the same successive swap transactions, or through a single swap transaction:
                $\Gamma \xrightarrow{swap(d\tau_\text{In}+d\tau_\text{In}^\prime)} \Gamma_1$.
    \end{enumerate}
\end{property}
\begin{proof}
    \hfill \linebreak
    We have two consecutive output slippage transactions on $\Gamma \xrightarrow{g(\Delta\tau_\text{Out}, \Gamma)} \Gamma^\prime \xrightarrow{g(\Delta\tau_\text{Out}^\prime, \Gamma^\prime)} \Gamma^{\prime\prime}$.\\
    \begin{enumerate}
        \item Let $g(\Delta\tau_\text{Out}; \Gamma)$ be the first swap transaction.
    \begin{align*}
        g(\Delta\tau_\text{Out}; \Gamma)
        &= \frac{R\tau_\text{Out}}{X\tau_\text{Out} + \Delta\tau_\text{Out}}\Delta\tau_\text{Out}\\
        &= \underbrace{\frac{R\tau_\text{Out}^2}{TB^2 + \Delta\tau_\text{Out} R\tau_\text{Out}}}_{\alpha}\Delta\tau_\text{Out}\\
        &= R\tau_\text{Out} - \frac{R\tau_\text{Out}T^2}{T^2+R\tau_\text{Out}\Delta\tau_\text{Out}}.
    \end{align*}
    \item Let $g(\Delta\tau_\text{Out}^\prime; \Gamma^\prime)$ be the second swap transaction followed by the first one.
    \begin{align*}
        g(\Delta\tau_\text{Out}^\prime; \Gamma^\prime) 
        &= R\tau_\text{Out}^\prime - \frac{R\tau_\text{Out}^\prime TB^2}{TB^2+(R\tau_\text{Out}^\prime)+\Delta\tau_\text{Out}^\prime}\\
        &= \frac{R\tau_\text{Out}T^2}{T^2+R\tau_\text{Out}\Delta\tau_\text{Out}} - \frac{TB^2\cdot(R\tau_\text{Out}-\alpha \cdot \Delta\tau_\text{Out}) }{TB^2+(R\tau_\text{Out}^\prime - \alpha\cdot\Delta\tau_\text{Out} )\Delta\tau_\text{Out}^\prime}\\
    \end{align*}
    \item Combine the output tokens of the two transactions above:
    \begin{align*}
        g(\Delta\tau_\text{Out}; \Gamma) + g(\Delta\tau_\text{Out}^\prime; \Gamma^\prime) 
            &= R\tau_\text{Out} -\frac{TB^2\cdot(R\tau_\text{Out}-\alpha \cdot \Delta\tau_\text{Out}) }{TB^2+(R\tau_\text{Out}^\prime - \alpha\cdot\Delta\tau_\text{Out} )\Delta\tau_\text{Out}^\prime}\\
        g(\Delta\tau_\text{Out}+\Delta\tau_\text{Out}^\prime; \Gamma) &= R\tau_\text{Out} - \frac{R\tau_\text{Out}T^2}{T^2+R\tau_\text{Out}(\Delta\tau_\text{Out}+\Delta\tau_\text{Out}^\prime)}
    \end{align*}
    \item Let $g(\Delta\tau_\text{Out}+\Delta\tau_\text{Out}^\prime; \Gamma)$ be a single transaction, that is separate from the above two successive transactions.
    \begin{align*}
        g(\Delta\tau_\text{Out}+\Delta\tau_\text{Out}^\prime; \Gamma) &= R\tau_\text{Out} - \frac{R\tau_\text{Out}T^2}{T^2+R\tau_\text{Out}(\Delta\tau_\text{Out}+\Delta\tau_\text{Out}^\prime)}
    \end{align*}
    \end{enumerate}

    \noindent Now, we compare $g(\Delta\tau_\text{Out}; \Gamma) + g(\Delta\tau_\text{Out}^\prime; \Gamma^\prime)$ against $g(\Delta\tau_\text{Out}+\Delta\tau_\text{Out}^\prime; \Gamma)$
    \begin{align*}
        g(\Delta\tau_\text{Out}; \Gamma) &+ g(\Delta\tau_\text{Out}^\prime; \Gamma^\prime) - g(\Delta\tau_\text{Out}+\Delta\tau_\text{Out}^\prime; \Gamma) \\
            &= \left(R\tau_\text{Out} - \frac{TB^2\cdot R\tau_\text{Out}}{TB^2+(R\tau_\text{Out}^\prime - \alpha\cdot\Delta\tau_\text{Out} )\Delta\tau_\text{Out}^\prime}\right) 
                 - \left(R\tau_\text{Out} - \frac{TB^2R\tau_\text{Out}}{TB^2+R\tau_\text{Out}(\Delta\tau_\text{Out}+\Delta\tau_\text{Out}^\prime)}\right)\\
            &= \frac{R\tau_\text{Out}}{TB^2+R\tau_\text{Out}(R\tau_\text{Out}^\prime - \alpha\cdot\Delta\tau_\text{Out} )}
                - \frac{R\tau_\text{Out}}{TB^2+R\tau_\text{Out}(\Delta\tau_\text{Out}+\Delta\tau_\text{Out}^\prime)}\\
            &= \frac{1}{Z} \left[ (R\tau_\text{Out} - \alpha \cdot \Delta\tau_\text{Out}) \left(TB^2+R\tau_\text{Out} \cdot(\Delta\tau_\text{Out}+\Delta\tau_\text{Out}^\prime)\right) - R\left(TB^2+(R\tau_\text{Out} - \alpha \cdot \Delta\tau_\text{Out}) \Delta\tau_\text{Out}^\prime\right)\right] \\
            &= \frac{1}{Z} (R\tau_\text{Out}-\alpha\cdot \Delta\tau_\text{Out})(TB^2+R\tau_\text{Out} \Delta\tau_\text{Out}) - TB^2 \cdot R\tau_\text{Out} \\
            &= \frac{1}{Z}  \left(R\tau_\text{Out}^2 \Delta\tau_\text{Out} - TB^2\alpha\Delta\tau_\text{Out} - \alpha \Delta\tau_\text{Out}^2 R\tau_\text{Out} \right)\\
            &= \frac{1}{Z} \left(R\tau_\text{Out}^2 \Delta\tau_\text{Out} - \frac{TB^2 R\tau_\text{Out}^2 \Delta\tau_\text{Out}}{TB^2+R\tau_\text{Out}\Delta\tau_\text{Out}}
                        + \frac{R\tau_\text{Out}^3 \Delta\tau_\text{Out}^2}{TB^2+ \Delta\tau_\text{Out} R\tau_\text{Out}}\right)\\
            &= \frac{1}{Z} \left(\frac{R\tau_\text{Out}^2 \Delta\tau_\text{Out} (TB^2+R\tau_\text{Out}\Delta\tau_\text{Out}) - TB^2 R\tau_\text{Out}^2 \Delta\tau_\text{Out} + R\tau_\text{Out}^3 \Delta\tau_\text{Out}^2}{TB^2+R\tau_\text{Out}\Delta\tau_\text{Out}} \right) \\
            &= 0
    \end{align*}
    where $Z = \left(TB^2+R\tau_\text{Out}(R\tau_\text{Out}^\prime - \alpha\cdot\Delta\tau_\text{Out})\right)\left(TB^2+R\tau_\text{Out}(\Delta\tau_\text{Out}+\Delta\tau_\text{Out}^\prime)\right)$

\end{proof}

\begin{property}[Psuedo-Reversibility]
    $swap \circ swap(d\tau_\text{In}; \Gamma) \leq d\tau_\text{In}$ 
\end{property}
\begin{proof}
    Let $\tau_\text{Out} = \rho \cdot USX(d\tau_\text{In}; \Gamma) \cdot d\tau_\text{In}).$\\
    Case 1: when $TB < R\tau_\text{Out}$ and $TB < R\tau_\text{In}-d\tau_\text{Out}$, then $USX$ is reversible 
    \begin{align*}
        d\tau_\text{In}^\prime &= \frac{1}{\rho} \cdot USX( d\tau_\text{Out}; \Gamma) \cdot d\tau_\text{In}) \cdot d\tau_\text{Out}\\
                &= \frac{1}{\rho} \underbrace{\left( \rho \cdot d\tau_\text{In} \right)}_{d\tau_\text{Out}}\\
                & = d\tau_\text{In}.
    \end{align*}
    
    Case2: when when $TB > R\tau_\text{Out}$ and $TB < R\tau_\text{In}-d\tau_\text{Out}$,
    \begin{align*}
        d\tau_\text{In}^\prime &= \frac{1}{\rho} \cdot \underbrace{\left( \rho \cdot \underbrace{\frac{R\tau_\text{Out}}{X\tau_\text{Out}+\rho d\tau_\text{In}}}_{<1} \cdot d\tau_\text{In}\right)}_{d\tau\text{Out}}
                < d\tau_\text{In}
    \end{align*}
    
    Case3: when when $TB > R\tau_\text{Out}$ and $TB > R\tau_\text{In}+d\tau_\text{In}$,
        \begin{align*}
        d\tau_\text{In}^\prime &= \frac{1}{\rho} 
            \cdot \underbrace{\left( \frac{R\tau_\text{In}+ d\tau_\text{In}}{X\tau_\text{In} + \rho \cdot \alpha d\tau_\text{In}}\right)}_{<1} 
            \cdot \left( \rho \cdot \underbrace{\frac{R\tau_\text{Out}}{X\tau_\text{Out}+\rho d\tau_\text{In}}}_{<1} \cdot d\tau_\text{In}\right) 
                < d\tau_\text{In}
    \end{align*}
    Case4: when when $R\tau_\text{Out} > TB > R\tau_\text{Out}-d\tau_\text{Out} $ and $TB > R\tau_\text{In}+d\tau_\text{In}$ and\\
    Case5: when when $R\tau_\text{Out} > TB > R\tau_\text{Out}-d\tau_\text{Out} $ and $R\tau_\text{In}+d\tau_\text{In} > TB$ will be the same as above.
\end{proof}

\subsubsection{UAMM swap transaction including input slippage function}
\label{app:uamm_swap_input_usx}
\label{app:slippage_funcs}

Here are the complete versions of input and output slippage functions that are continuous with respect to the inputs $d\tau_\text{In}$ and $\Delta\tau_\text{Out}$,
\begin{align*}
    d\tau_{\text{Out}} &=  \underbrace{g \Big( \underbrace{ \rho  \underbrace{f (d\tau_{\text{In}}; \Gamma)}_{\text{\scriptsize Input Slippage }\Delta\tau_{\text{In}}}}_{\Delta\tau_{\text{Out}}=\rho \Delta\tau_{\text{In}}} ; \Gamma \Big)}_{\text{Output Slippage } d\tau_{\text{Out}}}.
\end{align*}
where $\Gamma$ is the state parameters and $\rho = \frac{\rho\tau_{\text{In}}}{\rho\tau_{\text{Out}}}$ is the price ratio.
Here is the following steps that to compute $d\tau_{\text{Out}}$:
\begin{enumerate}
    \item compute the slippage on input pool $R\tau_{\text{In}}$, $\Delta\tau_{\text{In}}=f (d\tau_{\text{In}}; \Gamma)$, 
    \item compute the swapping amount based on price ratio, $\Delta\tau_{\text{Out}} = \rho \cdot \Delta\tau_{\text{In}}$. 
    \item compute the slippage on output pool $R\tau_{\text{Out}}$, $d\tau_{\text{Out}} = g(\Delta\tau_{\text{Out}}; \Gamma)$.
\end{enumerate}
\begin{align*}
    f(d\tau_{\text{In}}; \Gamma) =
        \begin{cases} 
            (\delta -1) (TB - R\tau_\text{In}) + d\tau_\text{In}  & \text{ if } R\tau_{\text{In}} \leq TB < R\tau_{\text{In}}+ d\tau_{\text{In}}\\
            d\tau_{\text{In}} & \text{ if } TB \geq R\tau_{\text{In}} \\
            d\tau_{\text{In}}\cdot \delta & \text{ otherwise }
        \end{cases}
\end{align*}
where $\delta = \frac{ R_{\text{In}} +d\tau_{\text{In}}}{X\tau_{\text{In}}}$.
The output slippage function $g(\Delta\tau_{\text{In}}; \Gamma)$ is already defined in the main paper.
The first if statements in $f$ are to ensure that the functions are continuous with respect to $d\tau_{\text{In}}$ and $\Delta\tau_{\text{Out}}$ respectively.

Based on our target balance $TB$ and current liquidity pool $R\tau_{\text{In}}$, we define a constant product curve $TB^2=R\tau_{\text{In}}X\tau_{\text{In}}$.
Then, $X\tau_{\text{In}}$ becomes the virtual pool that we use to encourage the liquidity pool balance $R\tau_{\text{In}}$ towards the target balance $TB$. 
For example, adding $d\tau_{\text{In}}$ into $R\tau_{\text{In}}$ would decrease the virtual pool balance $X\tau_{\text{In}}$ in order to satisfy the constant product curve,
\begin{align*}
    TB^2=(R\tau_{\text{In}}+d\tau_{\text{In}})(X\tau_{\text{In}}-\text{CP}(d\tau_{\text{In}})).
\end{align*}
Re-arranging the above equation, we get 
\begin{align*}
    \text{CP}(d\tau_{\text{In}}) = X\tau_{\text{In}} - \frac{TB^2}{R\tau_{\text{In}}+d\tau_{\text{In}}} =  d\tau_{\text{In}} \frac{X\tau_{\text{In}}}{R\tau_{\text{In}}+d\tau_{\text{In}}} = d\tau_{\text{In}} \underbrace{\frac{TB}{R\tau_\text{In}}}_{\geq 1}\underbrace{\frac{TB}{R\tau_{\text{In}}+d\tau_{\text{In}}}}_{\geq 1}.
\end{align*}
This corresponds to swapping $d\tau_{\text{In}}$ for $\text{CP}(d\tau_{\text{In}})$.
The last equality illustrates that $\text{CP}(d\tau_{\text{In}}) \geq d\tau_{\text{In}}$. 

The input slippage is defined as the relative difference between input token amount $d\tau_{\text{In}}$ and the swapped token amount $\text{CP}(d\tau_{\text{In}})$,
and finally, the input slippage function becomes
\begin{align*}
    \Delta_{\text{In}} = f(d\tau_{\text{In}}; \theta) = d\tau_{\text{In}} \underbrace{\frac{d\tau_{\text{In}}}{\text{CP}(d\tau_{\text{In}})}}_{0 < \text{slippage} \leq 1} =  d\tau_{\text{In}} \frac{R\tau_{\text{In}}+d\tau_{\text{In}}}{X\tau_{\text{In}}}.
\end{align*}

\begin{definition}[UBET slippage rate including input slippage]  
    \label{def:full_ubet_slippage}
    The UBET Automated Market Maker rate function is
    \begin{align}
        \label{eqn:ubet_slippage_rate_full}
        USX(d\tau_{\text{In}}, \Gamma) = 
        \begin{cases}
            \eta & \text{ if $TB \leq R\tau_{\text{Out}}- \Delta\tau_{\text{Out}} $ and $R\tau_{\text{In}} < TB < R\tau_{\text{In}}+d\tau_{\text{In}}$ }\\
            \eta \cdot \xi(\eta) & \text{ if $R\tau_{\text{Out}} - \Delta\tau_{\text{Out}} < TB < R\tau_{\text{Out}}$ and $R\tau_{\text{In}} < TB < R\tau_{\text{In}}+d\tau_{\text{In}}$ }\\
            \xi(\Delta\tau_{\text{Out}})  & \text{ if $R\tau_{\text{Out}} - \Delta\tau_{\text{Out}} < TB < R\tau_{\text{Out}}$ and $TB \leq R\tau_{\text{In}}$ }\\
            1 & \text{ if $TB \leq R\tau_{\text{Out}}- \Delta\tau_{\text{Out}}$ and $TB \geq R\tau_{\text{In}}$ }\\
            \frac{\Delta\tau_{\text{In}}}{d\tau_{\text{In}}} & \text{ if }  TB \leq R\tau_{\text{Out}}- \Delta\tau_{\text{Out}} \text{ and } TB < R\tau_{\text{In}} \\
            \alpha & \text{ if }  TB > R\tau_{\text{Out}} \text{ and } TB \geq R\tau_{\text{In}}\\
            \frac{\Delta\tau_{\text{In}}}{d\tau_{\text{In}}} \cdot \alpha  & \text{ otherwise }
        \end{cases}
    \end{align}
    where 
    \begin{align*}
        \alpha &= \frac{R\tau_{\text{Out}}}{X\tau_{\text{Out}}+ \Delta\tau_{\text{Out}}}\\
        \delta &= \frac{ R_{\text{In}} +d\tau_{\text{In}}}{X\tau_{\text{In}}}\\
        \eta &= (\delta -1) (TB - R\tau_\text{In}) + d\tau_\text{In}\\
        \xi(x) &=  \left(R\tau_\text{Out}-TB\right)\rho + \left(x - \left(R\tau_\text{Out}-TB\right)\right) \frac{R\tau_{\text{Out}}}{X\tau_{\text{Out}}+\rho x} 
    \end{align*}
    where $R\tau_{\text{Out}} - \Delta\tau_\text{Out}< TB$ cases ensure that we add output slippage if the swap amount with (or without) input slippage does not exceed the output liquidity pool amount, $\Delta\tau_\text{Out} \geq R\tau_{\text{Out}}$.
\end{definition}

This UAMM slippage does satisfy {\em open-boundedness} and {\em homogeneity} but does not satisfy {\em monotonicity} and {\em additivity} constraints. 
However, this does not mean that this UAMM with input slippage is invalid. 
It just generates different properties. For example, it will encourage that LPs to have zero impermanent loss and gain, whereas
 the one without input slippage will be generous and allow LPs to have impermanent gain.

\subsection{Spontaneous Price Derivation}

The spontaneous price is our market price with no slippage. 
One  can consider the spontaneous price as a fair price with an applied spread.
Because the price depends on whether $R\tau_{\text{In}}$ and $R\tau_{\text{Out}}$ are below the target price or not,
the spontaneous price is also a piece-wise function, 
\begin{align}
    \rho = \frac{p\tau_\text{In}}{p\tau_\text{Out}} = \left(\frac{\text{d} d\tau_{\text{In}}} {\text{d} d\tau_{\text{Out}}}\Bigg|_{d\tau_{\text{Out}}=0}\right)^{-1}
\end{align}
where
\begin{align*}
    \frac{\text{d} d\tau_{\text{In}}} {\text{d} d\tau_{\text{Out}}}\Bigg|_{d\tau_{\text{Out}}=0} =
        \begin{cases}
            \frac{1}{\rho} & \text{ if $TB \leq R\tau_{\text{In}}$ and $TB \leq R\tau_{\text{Out}}$  }\\
            \frac{X\tau_{\text{In}}}{\rho \cdot R\tau_{\text{In}}} & \text{ if }  TB \leq R\tau_{\text{Out}} \text{ and } TB < R\tau_{\text{In}} \\
            \frac{X\tau_{\text{Out}}}{\rho \cdot R\tau_{\text{Out}}} & \text{ if }  TB > R\tau_{\text{Out}} \text{ and } TB \leq R\tau_{\text{In}}\\
            \frac{X\tau_{\text{Out}}\cdot X\tau_{\text{In}}}{\rho \cdot R\tau_{\text{Out}} \cdot R\tau_{\text{In}}} & \text{ otherwise }.
        \end{cases}
\end{align*}
The marginal price of an output token in a constant product curve is defined as the price of an infinitesimally small trade. 

We differentiate the constant product curve with respect to the output token amount $d_{\text{Out}}$,
\begin{align}
    0 = \frac{\text{d}}{\text{d}d_{\text{Out}}} (R_{\text{Out}} - d_{\text{Out}}) (X_{\text{Out}}+ \rho \Delta_{\text{In}}) \bigg|_{d_{\text{Out}}=0} \Rightarrow
    \frac{\text{d}\Delta_{\text{In}}}{\text{d}d_{\text{Out}}} \bigg|_{d_{\text{Out}}=0} = \frac{X_{\text{Out}}+\rho \Delta_{\text{In}}}{\rho(R_{\text{Out}}-d_{\text{Out}})} \bigg|_{d_{\text{Out}}=0}
    \label{eqn:sp1}
\end{align}

We consider when $T-R_{\text{In}} >0 $, otherwise it is trivial.
\begin{align}
     \frac{\text{d}\Delta_{\text{In}}}{\text{d}d_{\text{Out}}} \bigg|_{d_{\text{Out}}=0} 
     = \frac{\text{d}}{\text{d}d_{\text{Out}}} \left( d_{\text{In}}\frac{R_\text{In}+d_{\text{In}}}{X_{\text{In}}}\right) \bigg|_{d_{\text{Out}}=0}
     = \frac{\text{d}d_{\text{In}}}{\text{d}d_{\text{Out}}}\frac{R_{\text{In}}+2d_{\text{In}}}{X_\text{In}} \bigg|_{d_{\text{Out}}=0}
    \label{eqn:sp2}
\end{align}

Substituting Equation~\ref{eqn:sp2} into Equation~\ref{eqn:sp1}, we get 
\begin{align*}
     \frac{X_{\text{Out}}+\rho \Delta_{\text{In}}}{\rho(R_{\text{Out}}-d_{\text{Out}})} \bigg|_{d_{\text{Out}}=0} &= \frac{\text{d}d_{\text{In}}}{\text{d}d_{\text{Out}}}\frac{R_{\text{In}}+2d_{\text{In}}}{X_\text{In}} \bigg|_{d_{\text{Out}}=0}\\
     \frac{\text{d}d_{\text{In}}}{\text{d}d_{\text{Out}}} &= \frac{X_{\text{Out}}X_{\text{In}}}{R_{\text{Out}}R_{\text{In}}\rho}.
\end{align*}